\def\year{2021}\relax
\documentclass[letterpaper]{article} 
\usepackage{aaai21}  
\usepackage{times}  
\usepackage{helvet} 
\usepackage{courier}  
\usepackage[hyphens]{url}  
\usepackage{graphicx} 
\usepackage{natbib}  
\usepackage{caption} 
\usepackage{microtype}
\usepackage{booktabs} 
\usepackage{subcaption}
\usepackage{enumitem}
\usepackage[final]{pdfpages}

\usepackage{amsmath,amsfonts,amssymb,amsthm}
\usepackage{commath}
\usepackage{mathtools}
\theoremstyle{definition}
\newtheorem{definition}{Definition}
\newtheorem{theorem}{Theorem}

\usepackage{pythonhighlight}

\title{Learning Discrete State Abstractions With Deep Variational Inference$\mathbb{^*}$}
\author{
    Ondrej Biza,
    Robert Platt$\mathbb{^{**}}$,
    Jan-Willem van de Meent$\mathbb{^{**}}$,
    Lawson L. S. Wong$\mathbb{^{**}}$
    \\
}
\affiliations{

    Khoury College of Computer Sciences, \\
    Northeastern University, Boston, MA, USA \\
    biza.o@northeastern.edu

}

\begin{document}

\maketitle

\begin{abstract}
Abstraction is crucial for effective sequential decision making in domains with large state spaces. In this work, we propose an information bottleneck method for learning approximate bisimulations, a type of state abstraction. We use a deep neural encoder to map states onto continuous embeddings. We map these embeddings onto a discrete representation using an action-conditioned hidden Markov model, which is trained end-to-end with the neural network. Our method is suited for environments with high-dimensional states and learns from a stream of experience collected by an agent acting in a Markov decision process. Through this learned discrete abstract model, we can efficiently plan for unseen goals in a multi-goal Reinforcement Learning setting. We test our method in simplified robotic manipulation domains with image states. We also compare it against previous model-based approaches to finding bisimulations in discrete grid-world-like environments. Source code is available at \url{https://github.com/ondrejba/discrete_abstractions}.
\end{abstract}

\section{Introduction}
\label{introduction}

High-dimensional state spaces are common in deep reinforcement learning \cite{mnih2015}. Although states may be as large as images, typically the information required to make good decisions is much smaller. This motivates the need for \emph{state abstraction}, the process of encoding states into compressed representations that retain features that inform action and discard uninformative features. 

\begin{figure}[t!]
    \centering
    \includegraphics[width=.8\linewidth]{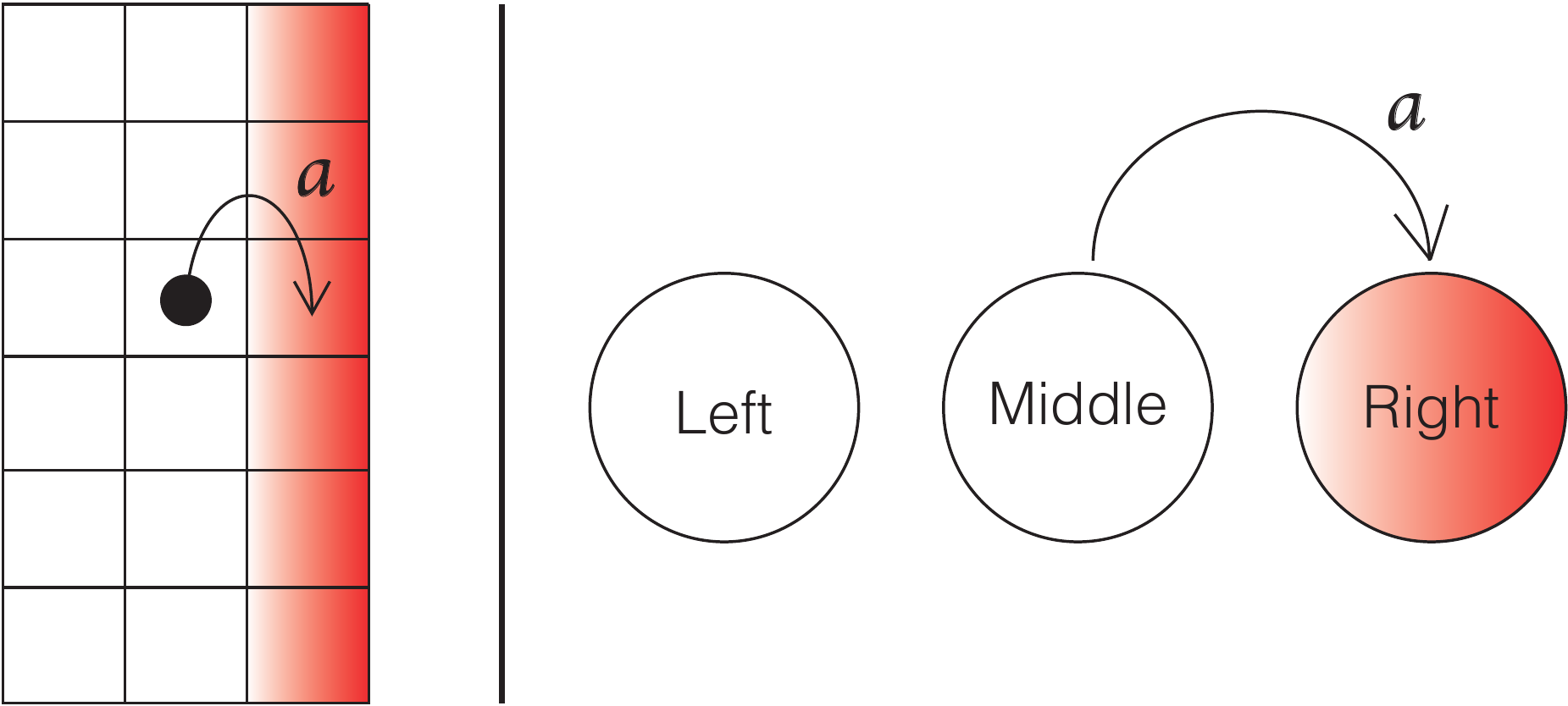}
    \caption{Example of bisimulation abstraction. The Column World (left) has 3 columns and 30 rows (we only show 6 rows); the agent travels between adjacent cells \cite{lehnert18}. Since the agent gets a reward 1 for being in the right column (red) and 0 otherwise, it is irrelevant in which row it is located. Hence, the environment can be simulated by an MDP with three states (right).}
    \label{fig_column_world}
\end{figure}

One principled approach to state abstraction is \emph{bisimulation} in Markov decision processes (MDP) \cite{dean97a}. Bisimulations formalize the notion of finding a smaller equivalent \emph{abstract} MDP that preserves transition and reward information, i.e., that retains relevant decision-making information while reducing the state space size. We demonstrate this idea in Figure \ref{fig_column_world}, where a grid world with fifteen states is compressed into an MDP with three states.


We pursue the bisimulation goal of finding a \emph{discrete abstract MDP} that can be used to plan policies. Unfortunately, finding bisimulations with maximally compressed state spaces is NP-hard \cite{dean97b}. One common approach to circumvent this is using bisimulation metrics, which facilitate transfer of \emph{existing} policies to similar states \cite{ferns04}. However, this method cannot generalize to \emph{new} tasks, for which there is no existing policy.




In this paper, we introduce an approach to finding approximate MDP bisimulations using the variational information bottleneck (VIB) \cite{tishby01,alemi17}. The VIB framework is typically used to learn representations that predict quantities of interest accurately while ignoring certain aspects of the domain.
VIB methods have previously been applied to state abstraction, but the learned abstraction does not in general take the form of an MDP bisimulation \cite{abel19}. This is problematic, because the abstract MDP can only represent the policies it was trained on, but cannot be used to plan for new tasks. To resolve this, whereas \citet{abel19} use the abstract states to predict \emph{actions} from an expert policy, we use abstract states to predict learned \emph{Q-values} in the VIB objective.


In our setup, a learned encoder maps a state $s$ in the original MDP into a continuous embedding $z$. We map the continuous state $z$ onto a discrete abstract state $\bar{s}$ by performing inference in a learned probabilistic model.
Our VIB method learns an encoder (i.e. a state abstraction $s \mapsto z$) that is predictive of the Q-values that are returned by a deep Q-network, but is regularized using structured prior over the embedding space ($z$). Concretely, we propose using priors that prefer clusters with Markovian transition structure. A sequence of embedded states $(z_1, z_2, \ldots)$ is treated as observations
from either a Gaussian mixture model (GMM) or an action-conditioned hidden Markov model (HMM),
where each embedding $z_t$ is emitted from a latent cluster representing abstract state $\bar{s}_t$.
In the HMM case, we also learn a cluster transition matrix for each action, serving as the abstract MDP transition model.
The key insight is that abstract states $\bar{s}$ group together ground states $s$ (and embeddings $z$) with similar Q-values and similar transition properties, thereby forming an approximate MDP bisimulation.

In addition to the neural encoder, the parameters of our GMM and HMM priors are learned as well. The learned parameters (cluster means, covariances, and discrete transition matrix between clusters) therefore form our abstract MDP state space and transition function. When presented with tasks not seen during training, we can use the learned abstract model to plan to solve these tasks without additional learning efficiently. 

In summary, our contributions are:
\begin{itemize}
\item Framing bisimulation learning as a VIB objective.
\item Introducing two structured priors (GMM, HMM) with learned parameters for VIB-based state abstraction.
\item Using the learned parameters of the prior to extract a discrete abstract MDP, which is an approximate bisimulation of the original MDP.
\item Using the abstract MDP to plan for new goals without requiring further training.
\end{itemize}

\section{Background}
\label{background}

\noindent
\textbf{Markov decision process:} We model our tasks as episodic Markov decision processes (MDPs). An MDP is a tuple $M = \langle S, A, T, R, \gamma \rangle$ \cite{bellman57}, where $S$ and $A$ are state and action sets, respectively. The function $R: S{\times}A \rightarrow \mathbb{R}$ describes the expected reward associated with each state-action pair. The density $T(s, a, s') = p(s' | s, a)$ describes transition probabilities between states. $\gamma \in \mathbb{R}$ is a discount factor. A policy $\pi: S{\times}A \rightarrow [0,1]$ encodes the behavior of an agent as a probability distribution over $A$ conditioned on $S$. The state-action value $Q_{\pi}$ of a policy $\pi$ is the expected discounted reward of executing action $a$ from state $s$ and subsequently following policy $\pi$:
\begin{align}
    Q_{\pi} (s, a) \coloneqq R(s,a) + \gamma \mathbb{E}_{s' \sim T,a' \sim \pi} \left[Q_{\pi} (s', a') \right].
\end{align}
We want to behave optimally both in the ground and abstract MDPs. A policy $\pi^*$ is optimal when $Q_{\pi^*}(s,a) \geq Q_\pi(s,a), \; \forall s,a \in S{\times}A$. 


\vspace{0.2cm}
\noindent
\textbf{State abstraction:} We approach state abstraction from the perspective of model minimization. The goal is to find a function that maps from the state space $S$ of the original MDP to a compact state space $\bar{S}$ while preserving the reward and transition dynamics \cite{dean97a,givan03}. Concretely, we want a surjective function $\phi : S \rightarrow \bar{S}$ that induces a partition over $S$. That is, each \emph{abstract state} $\bar{s} \in \bar{S}$ is associated with a block of states in $S$ defined by the preimage of $\phi$ at $\bar{s}$, $\phi^{-1}(\bar{s}) \subseteq S$. Since $\phi$ must induce a partition over $S$, we require $\phi^{-1}(\bar{s}_1) \cap \phi^{-1}(\bar{s}_2) = \emptyset, \; \forall \bar{s}_1,\bar{s}_2 \in \bar{S}$ such that $\bar{s}_1 \not= \bar{s}_2$. A \emph{bisimulation} is a surjection $\phi : S \rightarrow \bar{S}$ that induces a partition over $S$ and preserves the reward and transition dynamics. It is commonly formalized as:

\begin{definition}[MDP Bisimulation]\label{def_bisimulation}
Let $M = \langle S, A, T, R, \gamma \rangle$ and $\bar{M} = \langle \bar{S}, A, \bar{T}, \bar{R}, \gamma \rangle$ be MDPs. A function $\phi: S \rightarrow \bar{S}$ is an MDP bisimulation from $M$ to $\bar{M}$ if the preimage $\phi^{-1}$ induces a partition of $S$ and for each $s_1,s_2 \in S$, $\bar{s} \in \bar{S}$ and $a \in A$, $\phi(s_1) = \phi(s_2)$ implies both $R(s_1,a) = R(s_2,a)$ and $\sum_{s' \in \phi^{-1}(\bar{s})} T(s_1,a,s') = \sum_{s' \in \phi^{-1}(\bar{s})} T(s_2,a,s')$.
\end{definition}

Given two MDPs $M$ and $\bar{M}$, there may exist many bisimulations. These bisimulations can be placed in a partial order. Given two bisimulations $\phi$ and $\phi'$ from $M$ to $\bar{M}$, we will say that $\phi'$ is a \emph{refinement} of $\phi$ if the partition induced by $\phi'$ is a refinement of that induced by $\phi$. 

\vspace{0.2cm}
\noindent
\textbf{Information Bottleneck Methods:} We approach the state abstraction problem using deep information bottleneck (IB) methods~\cite{alemi17}. These methods assume that we provide a distribution $q(s,y)$ over features $s$ (in our case the state) and a prediction target $y$ (in our case expected reward or return). Given $q(s,y)$, we learn a neural encoder $q(z \mid s)$ that maps $s$ onto a compressed representation $z$ by maximizing the IB objective \cite{tishby01}
\begin{equation}
    \label{eq:info-bottleneck}
    R_{IB} = I(y;z) - \beta \; I(s;z).
\end{equation}
Here $I$ denotes the mutual information between its arguments. The intuition behind this objective is that we would like to learn a (lossy) compressed representation of $s$ by maximizing the correlation between $z$ and the target $y$, which ensures that $z$ is predictive of $y$, whilst minimizing the correlation between $z$ and $s$, which ensures that any information in $s$ that does not correlate with $y$ is discarded.

In practice, evaluating the IB objective is intractable. Instead of optimizing Equation \eqref{eq:info-bottleneck} directly, IB methods introduce two variational distributions $p(y \,|\, z)$ and $p(y)$ to bound the mutual information terms
\begin{align}
    I(y;z) 
    &\ge
    \mathbb{E}_{q(s,y,z)}
    \left[
    \log p(y | z)
    -
    \log q(y)
    \right]
    ,
    \\
    I(s;z) 
    &\le 
    \mathbb{E}_{q(s, y, z)}
    \left[
    \log q(z \,|\, s)
    - 
    \log p(z)
    \right].
\end{align}
Combining these terms bounds the IB objective %
\begin{align}
\nonumber
    R_{IB}
    &\geq
    \mathbb{E}_{q(s,y,z)}
    \left[
    \log p(y \mid z)
    +
    \beta 
    \log \frac{p(z)}{q(z \mid s)}
    \right]
    +
    H(y).
\end{align}
Maximizing the above with respect to $p(y \,|\, z)$, $p(z)$, and $q(z \mid s)$ is a form of variational expectation maximization; optimizing $p(y \,|\, z)$ and $p(z)$ tightens the bound, whereas optimizing $q(z \mid s)$ maximizes the IB objective. Note that the entropy term $H(y)$ does not depend on $q(z \,|\, s)$ and can therefore be ignored safely during optimization.

\section{Learning bisimulations}
\label{sec:learning-bismulations}

\begin{figure}[t!]
  \centering
  \includegraphics[width=\columnwidth]{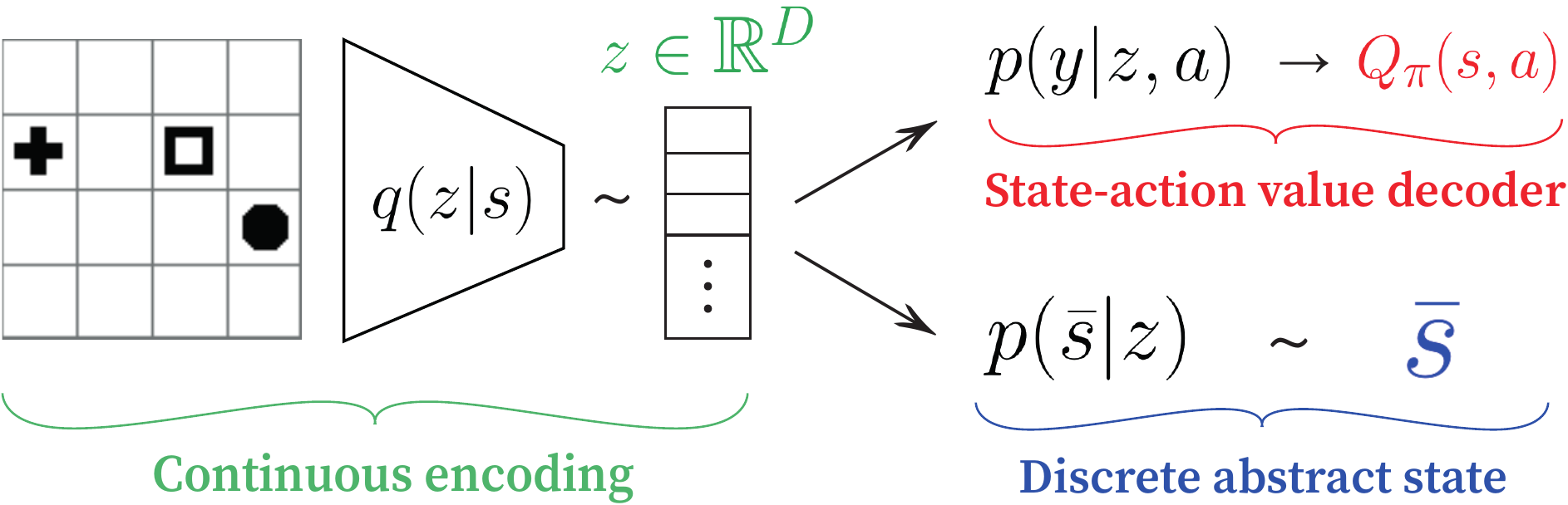}
  \caption{Inference in our model. We first take a state, represented as an image here, and encode it as a continuous vector $z$ (green). Then, we predict state-action values from $z$ for each action $a$ (red) and further encode $z$ into a discrete abstract state $\bar{s}$ using our prior (blue).}
  \label{fig_long_method}
\end{figure}

We propose a variational method for finding bisimulations directly from experience. The end result of our process is an abstract MDP, in which we can efficiently plan policies. Our model consists of three parts:
\begin{enumerate}
    \item A deep neural encoder $q(z \mid s)$ that projects states (usually represented as images) onto a low-dimensional continuous latent space (Figure \ref{fig_long_method} left).
    \item A generative model $p(z, \bar{s})$ with a tractable posterior \mbox{$p(\bar{s} \mid z)$}, that encodes the prior belief that the experience was generated by a small discrete Markov process (Figure \ref{fig_long_method} lower right).
    \item A linear decoder $p(y \mid z, a)$ that predicts state-action values from the continuous encodings (Figure \ref{fig_long_method} upper right).
\end{enumerate}

We tie the three models together using the deep variational information bottleneck method \cite{alemi17}. Unlike the standard setting, we encode pairs of ground states $(s_t, s_{t+1})$ as latent state pairs $(z_t, z_{t+1})$. This enables us to learn a tabular transition function between discrete states inside a prior $p(z_t, z_{t+1} | a)$. The discrete states of the prior together with the learned transition function and a given reward function define an abstract MDP. 

\subsection{Bisimulation as an information bottleneck}
\label{sub_methods_ib}

We apply deep IB methods to learn bisimulations as follows.
Let $q(s_t, a, y, s_{t+1})$ be a empirical distribution representing a dataset of transitions from state $s_t$ to state $s_{t+1}$ under an action $a$, selected by an arbitrary policy $\pi$. $y = Q_{\pi}(s_t,a)$ denotes the state-action value for the pair $(s_t,a)$ under $\pi$. We will use the IB method to find a compact latent encoding of $s_t$ that enables us to predict $y$ while simultaneously matching a prior on the temporal dynamics of the process that generated the data. Let $s = (s_t,s_{t+1})$ denote a sequential pair of states and $z = (z_t,z_{t+1})$ a corresponding sequential pair of latent states. The standard IB formulation is:
\begin{align}
\nonumber
    R_{IB} & = \mathbb{E}_{q(s,a,y,z)} \big[ I(y;z | a) - \beta I(s;z | a) \big] \\
\nonumber
    & \geq \mathbb{E}_{q(s,a,y,z)} \Big[ \log p(y|z,a) - \beta \log \frac{q(z|s,a)}{p(z|a)} \Big] \\
\nonumber
& \geq \mathbb{E}_{q(s,a,y,z)} \Big[ \log p(y|z_t,z_{t+1},a) - \\
\nonumber
& \qquad \qquad \qquad \qquad \beta \log \frac{q(z_t,z_{t+1}|s_t,s_{t+1},a)}{p(z_t,z_{t+1}|a)} \Big],
\end{align}
where use $q(s,a,y,z) = q(s_t, a, y, s_{t+1}) q(z|s,a)$ as shorthand notation and expand $s = (s_t,s_{t+1})$ and $z = (z_t,z_{t+1})$ in the last identity. 

We make two architectural decisions grounded in standard Markov assumptions. First, we assume that the value $y$ is conditionally independent of $z_{t+1}$ given $z_t$: $q(y|z_t,z_{t+1},a) = q(y|z_t,a)$. Second, we assume that $z_t$ is conditionally independent of $z_{t+1}$, $s_{t+1}$, and $a$ given $s_t$ and likewise that $z_{t+1}$ is conditionally independent of $z_t$, $s_t$, and $a$:
$q(z_t,z_{t+1}|s_t,s_{t+1},a) = q(z_t|s_t) q(z_{t+1}|s_{t+1})$.
Together, these two assumptions yield an IB lower bound of the form
\begin{align}
\nonumber
    L_{IB} 
    = 
    \mathbb{E}_{q(s,a,y,z)} 
    \Big[ 
        & \log p(y|z_t,a) 
    \\
        & 
        - \beta \log \frac{q(z_t|s_t)q(z_{t+1}|s_{t+1})}{p(z_t,z_{t+1}|a)} \Big].
    \label{eqn:theloss}
\end{align}
$L_{IB}$ presents a trade-off between encoding enough information of $s_t$ in order to predict $y$ (the first term of Equation~\ref{eqn:theloss}) and making the sequence $(z_t, z_{t+1})$ likely under our prior (the second term). This prior, $p(z_t,z_{t+1}|a)$, is a key element of our approach and is discussed in the next section.

Notice that $\log p(y \mid z_t, a)$ (the first term in Equation~\ref{eqn:theloss}) predicts a state-action value, not a reward. This provides additional supervision. Without it, the model tends to collapse by predicting a single abstract state for each ground state.

\subsection{Structured Priors}
\label{sub_methods_priors}

The denominator of the second term in Equation~\ref{eqn:theloss} $p(z_t,z_{t+1}|a)$ is the prior. We use this prior to incorporate an inductive bias, which is that we are observing a discrete Markov process. We evaluate two priors for this purpose: a prior based on a Gaussian mixture model and one based on an action-conditioned Hidden Markov model. 

\subsubsection{GMM Prior}

We assume $K$ components, each parameterized by a mean  $\mu_k$ and a covariance $\Sigma_k$
\begin{align}
p_\textsc{gmm} (z_t) & = \sum_{k=1}^K p(z_t | c_t = k) p(c_t = k) \\
    &= \sum_{k=1}^K \mathcal{N} (z_t | \mu_k, \Sigma_k) \rho_k.
\end{align}
Here $\rho_k = p(c_t = k)$ denotes the probability that $z_t$ was generated by component $k$ and $\mathcal{N} (z_t | \mu_k, \Sigma_k) = p(z_t | c_t = k)$ is the Gaussian distribution for the $k^{th}$ component. In this paper, we constrain $\Sigma_k$ to be diagonal. For the GMM prior, we set $p(z_t,z_{t+1}|a) = p_\textsc{gmm} (z_t)$ and allow $\mu_k$ and $\Sigma_k$ to vary; $\rho_k$ is uniform and fixed. This encodes a desire to find a latent encoding generated by membership in a finite set of discrete states (the mixture components). Each mixture component corresponds to a distinct abstract state. The weighting function $\rho_k$ is the probability that the continuous encoding $z_t$ was generated by the $k^{th}$ abstract state. This encodes the prior belief that latent encoding of state should be distributed according to a mixture of Gaussians with unknown mean, covariance, and weights. Note that while this approach gives us an encoder that projects real-valued high dimensional states onto a small discrete set of abstract states, it ignores the temporal aspect of the Markov process.

\subsubsection{HMM Prior}

\begin{figure}
    \centering
    \includegraphics[width=0.8\linewidth]{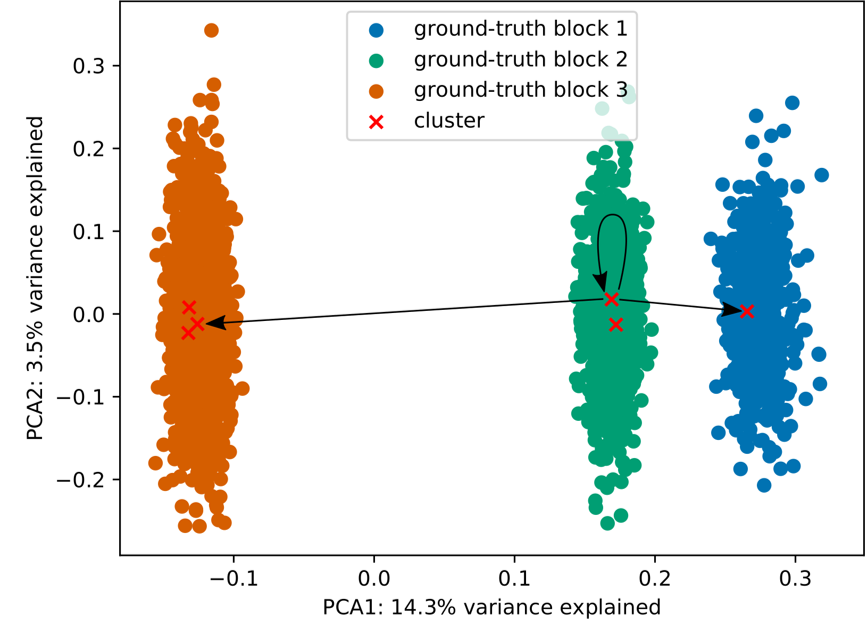}
    \caption{Latent states in the Column World environment found using our model (PCA projection). The colors represent the three blocks in the coarsest bisimulation (see Figure \ref{fig_column_world} left), which are unknown during training. The red crosses denote the locations of the six component means and the arrows illustrate the transition function found by the HMM prior.}
    \label{figx}
\end{figure}

To capture the temporal aspect of a Markov process, we can model the prior as an action conditioned hidden Markov model (an HMM). Here, the ``hidden'' state is the unobserved discrete abstract state $c_t$ used to generate ``observations'' of the latent state $z_t$. As in the GMM, there are $K$ discrete abstract states, each of which generates latent states according to a multivariate Normal distribution with mean $\mu_k$ and (diagonal) covariance matrix $\Sigma_k$. Since we are modelling a Markov process, we include a separate transition matrix $T^a$ for each action $a$ where $T^a_{k,l}$ denotes the probability of transitioning from an abstract state $k$ to an abstract state $l$ under an action $a$. Using this model, the prior becomes:
\begin{align}
    &p_\textsc{hmm} (z_t, z_{t+1} | a) = \sum_{k=1}^K \sum_{l=1}^K p(z_t | c_t = k) \\
    & \quad p(z_{t+1} | c_{t+1} = l) p(c_{t+1} = l | c_t = k, a) p (c_t = k) \\
    &= \sum_{k=1}^K \sum_{l=1}^K \mathcal{N} (z_t | \mu_k, \Sigma_k) \mathcal{N} (z_{t+1} | \mu_l, \Sigma_l) T^a_{k,l} \rho_k
\end{align}
As with the GMM prior, we allow the parameters of this model ($\mu_k$, $\Sigma_k$, and $T^a$) to vary, except for $\rho_k$, which is uniform and fixed. The transition model $T^a$ found during optimization is a discrete conditional probability table that defines a discrete abstract MDP. Essentially, this method finds the parameters of a hidden discrete abstract MDP that fits the observed data over which the loss of Equation~\ref{eqn:theloss} is evaluated.

Figure~\ref{figx} illustrates the latent embedding found using the HMM model for the Column World domain shown in Figure~\ref{fig_column_world}. The three clusters correspond to the three states in the coarsest abstract bisimulation MDP. These three clusters are overrepresented by six cluster centroids (the red x's) because we ran our algorithm using six cluster components. The algorithm ``shared'' these six mixture components among the three bisimulation classes. The result is still a bisimulation--just not the coarsest bisimulation.

\subsection{Deep encoder and end-to-end training}
\label{sub_methods_nn}

The loss $L_{IB}$ (Equation~\ref{eqn:theloss}) is defined in terms of three distributions that we need to parameterize: the encoder $q(z|s)$, the Q-predictor $p(y|z_t,a)$ and the prior $p(z_t,z_{t+1}|a)$. The encoder $q(z|s)$ is a convolutional network that predicts the mean $\mu^{CNN} (s_t)$ and the diagonal covariance $\Sigma^{CNN} (s_t)$ of a multivariate normal distribution. We used a modified version of the encoder architecture from \cite{ha18}\footnote{Appendix A.1. in version 4 of their arXiv submission.}: five convolutional layers followed by one fully-connected hidden layers and two fully-connected heads for $\mu^{CNN}$ and $\Sigma^{CNN}$, respectively. The Q-predictor $p(y|z_t,a)$ is a single fully-connected layer (i.e. a linear transform). We chose this parameterization to impose another constraint on the latent space: the encodings $z$ not only need to form clearly separable clusters to adhere to the prior, but also linearly dependent on their state-action values for each action. When we train on state-action values for multiple tasks, we predict a vector $y$ instead of a scalar. Using the reparameterization trick to sample from $q(z|s)$, we can compute the gradient of the objective with respect to the encoder weights, Q-predictor weights and the prior parameters. The prior parameters include the component means and variance, together with the transition function for hidden states in the HMM.

\subsection{Planning in the Abstract MDP}
\label{sub_abstract MDP}

A key aspect of our approach is that we can solve new tasks in the original problem domain by solving a compact discrete MDP for new policies. This is one of the critical motivations for using bisimulations: optimal policies in the abstract MDP induce optimal policies in the original MDP. We define the abstract MDP $\bar{M} = \langle \bar{S}, A, \bar{T}, \bar{R}, \gamma \rangle$ with the discrete transition table learned by the HMM prior. The abstract reward function can encode any reward function in the ground MDP by projecting ground rewards into the abstract space using the encoder. Now, we can use standard discrete value iteration to find new policies. These policies can be immediately applied in the ground MDP: observations of state in the ground MDP can be projected into the discrete abstract MDP and the new policy can be used to calculate an action. 




\section{From VIB abstraction to bisimulation}
\label{vib_and_bisim}

The HMM embedded in our model learns parameters of an compact discrete MDP (Subsection \ref{sub_methods_priors}), but it is not a priori clear that this abstraction is also a bisimulation. We show that under idealized conditions, every optimal solution to the objective $L_{IB}$ is a bisimulation. We analyze the idealized case where the following assumptions hold:
\begin{enumerate}
    \item The transitions in the ground MDP are deterministic;
    \item The HMM prior has enough components to represent a bisimulation, no two components share a mean;
    \item The prior over hidden states $p(c_t = k)$ is fixed;
    \item The encoder is deterministic and the prior observation model is an identity function over component means;
    \item The decoder $p(y|z_t,a)$ makes a prediction for each component using a table of state-action values.
\end{enumerate}

\begin{theorem}
\label{theorem_hmm_bisim}

There exist model parameters $\theta$ that reach the global minimum of $L_{IB}(\theta) = \beta \log K$ with $\beta > 0$. The abstraction mapping $\phi$ induced by any such model parameters is a bisimulation.

\end{theorem}

See Section A in the Appendix for the proof.

\section{Experiments}

The aim of our experiments is to investigate the following aspects of our method:
\begin{itemize}
    \item its ability to find abstractions that are compact and accurately model the ground MDP,
    \item planning in the abstract MDP for new goals, and
    \item its performance in environments without a clear notion of an abstract state.
\end{itemize}

We compare our method against an approximate bisimulation baseline (Section \ref{exp_columns}) in a grid world and test it in more complex domains with image states (Section \ref{exp_shapes} and Appendix C.1) and in simplified Atari games (Section \ref{exp_minatar}).



\subsection{Column World}
\label{exp_columns}

The purpose of this experiment is to compare our method to a model-based approximate bisimulation baseline in a simple discrete environment. Column World is a grid world with 30 rows and 3 columns \cite{lehnert18}. The agent can move left, right, top and down, and it receives a reward 1 for any action executed in the right column; otherwise, it gets 0 reward. Hence, the agent only needs to know if it is in the left, middle or right column, as illustrated in Figure \ref{fig_column_world}.


First, we train a deep Q-network on this task and use it to generate a dataset of transitions. As a baseline, we train a neural network model to predict $r_t$ and $s_{t+1}$ given $(s_t,a_t)$. We then find a coarse approximate bisimulation for this model using a greedy algorithm from \cite{dean97b} with the approximation constant $\epsilon$ set to 0.5. We compare it with our method trained with an HMM prior on $Q_{\pi}(s_t,a_t)$ predicted by the deep Q-network. We represent each state as a discrete symbol and use fully-connected neural networks for all of our models. See Appendix B.2 for details.

\begin{figure}[!t]
    \centering
    \begin{subfigure}[b]{0.7\linewidth}    
         \includegraphics[width=1.0\textwidth]{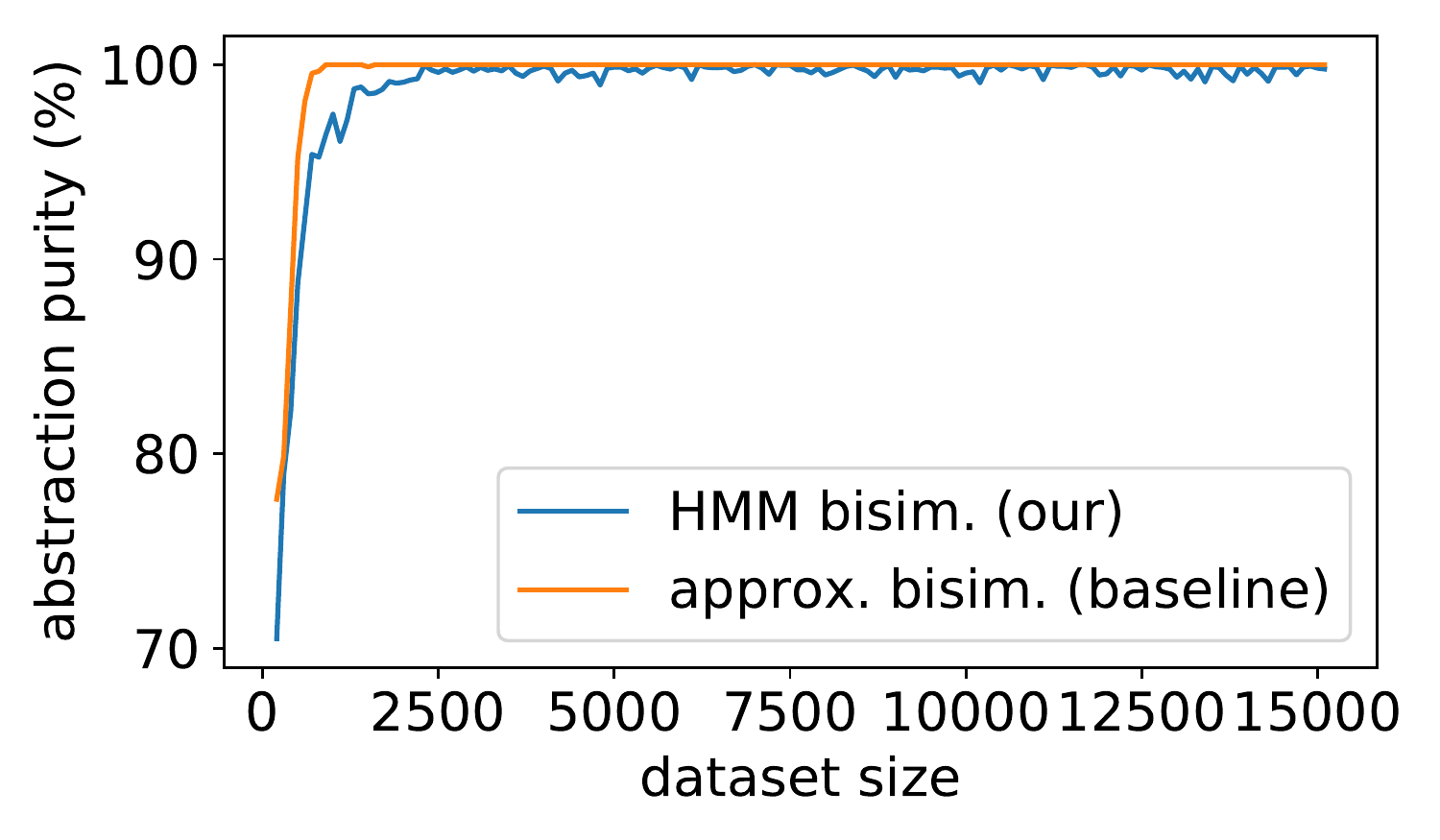}
     \end{subfigure} 
     \begin{subfigure}[b]{0.7\linewidth}    
         \includegraphics[width=1.0\textwidth]{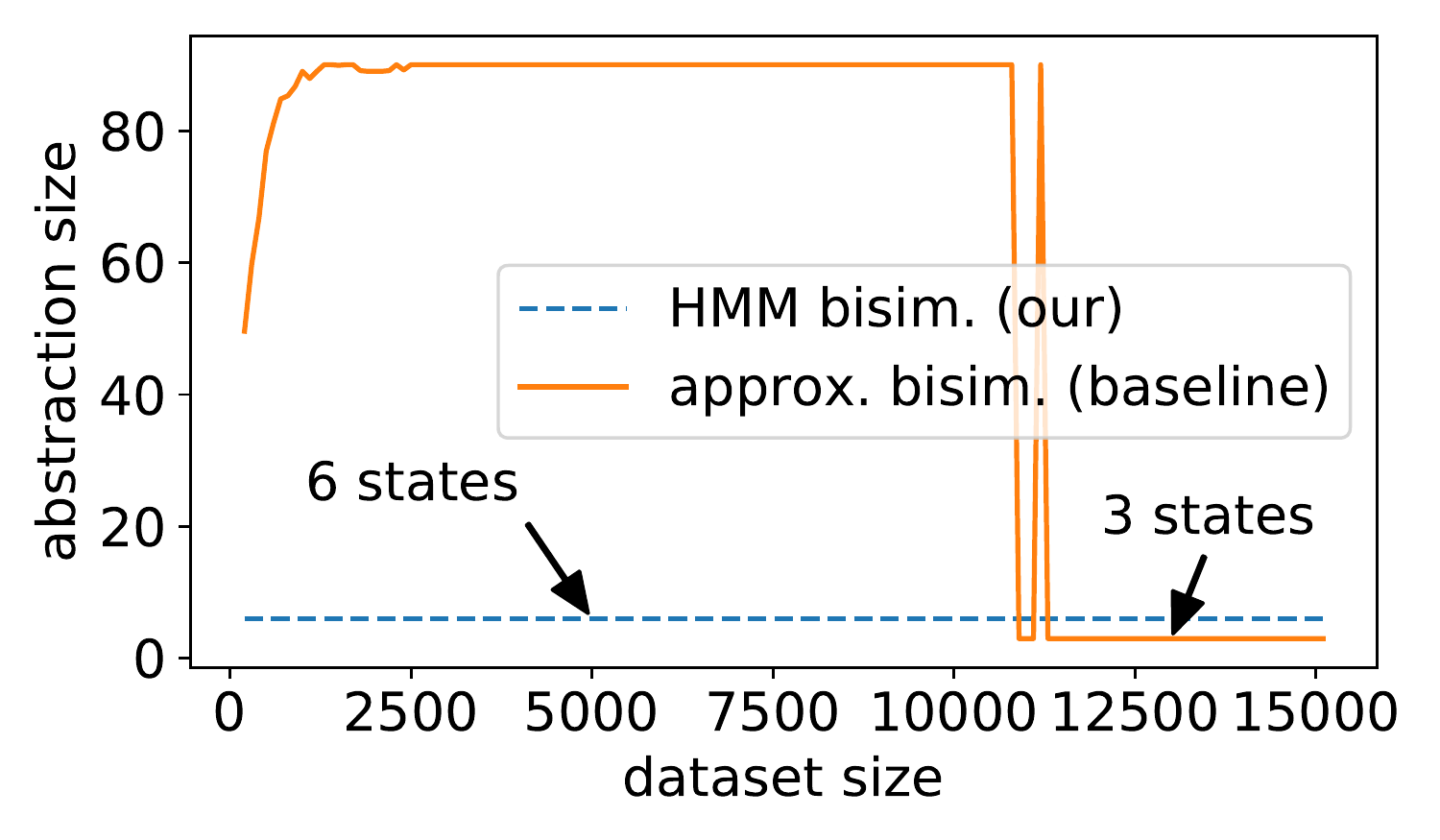}
     \end{subfigure}
    \caption{Comparison between model-based approximate bisimulation and our method in Column World \cite{lehnert18}. We vary the dataset sizes used for learning the bisimulation from 1000 to 20000 samples. Abstraction size refers to the number of abstract states.}
    \label{fig_res_column_world}
\end{figure}

Figure \ref{fig_res_column_world} shows the purity and the size of the abstractions found by our method and the baseline as a function of dataset size. We need a ground-truth abstraction to calculate the abstraction purity--in this case, it is the three-state abstraction shown in Figure \ref{fig_column_world} right. We assign each ground state to an abstract state (Figure \ref{fig_long_method}) and find the most common ground-truth label for each abstract state. The abstraction purity is the weighted average of the fraction of members of an abstract states that share its label. We include a snippet of code that computes purity in Appendix B.1. 

Both methods can find an abstraction with high purity. However, approximate bisimulation does not reduce the state space (there are 90 ground states) until the model of the environment is nearly perfect, which requires more than 11000 training examples. Our method always finds an abstraction with six states (the number of abstract states is a hyper-parameter), but our method finds a compact high-purity abstraction much faster than the baseline. Notice that we parameterize our method with more abstract states than the size of the coarsest bisimulation. In practice, this over-parameterization aids convergence.

\subsection{Shapes World}
\label{exp_shapes}

\begin{figure*}[t!]
    \centering
    \begin{subfigure}[b]{0.12\textwidth}    
         \includegraphics[width=1.0\textwidth]{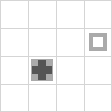}
     \end{subfigure} 
     \begin{subfigure}[b]{0.12\textwidth}    
         \includegraphics[width=1.0\textwidth]{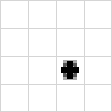}
     \end{subfigure}
     \begin{subfigure}[b]{0.12\textwidth}    
         \includegraphics[width=1.0\textwidth]{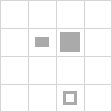}
     \end{subfigure}
     \begin{subfigure}[b]{0.12\textwidth}    
         \includegraphics[width=1.0\textwidth]{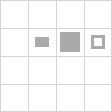}
     \end{subfigure}
     \begin{subfigure}[b]{0.12\textwidth}    
         \includegraphics[width=1.0\textwidth]{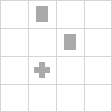}
     \end{subfigure}
     \begin{subfigure}[b]{0.12\textwidth}    
         \includegraphics[width=1.0\textwidth]{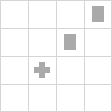}
     \end{subfigure}
     \begin{subfigure}[b]{0.12\textwidth}    
         \includegraphics[width=1.0\textwidth]{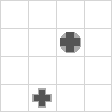}
     \end{subfigure}
     \begin{subfigure}[b]{0.12\textwidth}    
         \includegraphics[width=1.0\textwidth]{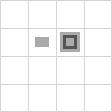}
     \end{subfigure}
    \caption{Goal states for tasks in Shapes World. There are four types of objects--pucks, boxes, squares and pluses--placed in a grid world. From left to right we have examples of goal states for two objects stacking, three objects stacking, two objects in a row, three objects in a row, two objects diagonal, three objects diagonal, two and two objects stacking, stairs from three objects.}
    \label{fig_goals}
\end{figure*}

We use a modified version of the Pucks World from \citet{biza19}. The world is divided into a $4{\times}4$ grid and objects of various shapes and sizes can be placed or stacked in each cell. States are represented as simulated $64{\times}64$ depth images. The agent can execute a PICK or PLACE action in each of the cells to pick up and place objects. The goal of abstraction is to recognize that the shapes and sizes of objects do not influence the PICK and PLACE actions. We instantiate eight different tasks in this domain described in Figure \ref{fig_goals}.

First, we test the ability of our algorithm to find accurate bisimulation partitions. Table \ref{tab_gmm_hmm} shows the results for our method for both the GMM and the HMM prior. 
Both models reach a high abstraction purity (described in Section \ref{exp_columns}) in all cases except for the three objects stacking task in a $4{\times}4$ grid world. The smallest MDP for which a bisimulation exists contains 936 abstract states; our algorithm has 1000 possible abstract states available. Our experiment shows that the HMM prior can leverage the temporal information, which is missing from the GMM, to allocate abstract states better.

\begin{table}[t!]
    \centering
    \scalebox{1}{
    \begin{tabular}{ccc}
        \toprule
        \textbf{Setting} & \textbf{GMM} & \textbf{HMM} \\
        \midrule
        2 pucks, $2{\times}2$ grid & $100$ \small{$\pm 0$} & $97$ \small{$\pm 1$} \\
        2 pucks, $3{\times}3$ grid & $100$ \small{$\pm 0$} & $98$ \small{$\pm 1$} \\
        2 pucks, $4{\times}4$ grid & $99$ \small{$\pm 1$} & $97$ \small{$\pm 0$} \\
        3 pucks, $2{\times}2$ grid & $99$ \small{$\pm 1$} & $97$ \small{$\pm 1$} \\
        3 pucks, $3{\times}3$ grid & $99$ \small{$\pm 1$} & $96$ \small{$\pm 1$} \\
        3 pucks, $4{\times}4$ grid & $56$ \small{$\pm 15$} & $89$ \small{$\pm 1$} \\
        \hline
        2 objects, $2{\times}2$ grid & $100$ \small{$\pm 0$} & $97$ \small{$\pm 1$} \\
        2 objects, $3{\times}3$ grid & $100$ \small{$\pm 0$} & $97$ \small{$\pm 0$} \\
        2 objects, $4{\times}4$ grid & $100$ \small{$\pm 0$} & $97$ \small{$\pm 0$} \\
        3 objects, $2{\times}2$ grid & $100$ \small{$\pm 0$} & $97$ \small{$\pm 1$} \\
        3 objects, $3{\times}3$ grid & $98$ \small{$\pm 1$} & $96$ \small{$\pm 1$} \\
        3 objects, $4{\times}4$ grid & $67$ \small{$\pm 3$} & $91$ \small{$\pm 1$} \\
        \bottomrule
    \end{tabular}
    }
    \caption{Results for learning abstractions for puck stacking and stacking of objects of various shapes. The difference between the top and bottom section is that the top section involves manipulating objects of only one type (pucks), whereas the bottom section involves four object types (puck, box, square and plus; see Figure \ref{fig_goals}). GMM and HMM refer to the two types of priors our model uses (Subsection \ref{sub_methods_priors}) and we report \textit{abstraction purities (\%)}. We report means and standard deviations over 10 runs.} 
    \label{tab_gmm_hmm}
\end{table}

Next, we test the ability of the learned abstract models to plan for new goals. We are able to reach a goal only if it is represented as a distinct abstract state in our model--such abstract states can only exist if the training dataset contains examples of the goal. Therefore, we can generalize to unseen goals in the sense that our model does not know about these goals during training, but they are represented in the dataset. During planning for a particular goal, we create a new reward function for the abstract model and assign a reward 1 to all transitions in the dataset that reach that goal. Then, we run Value Iteration in the abstract model and use the found state-action values to create a stochastic softmax policy. See Appendix B.3 for more details. 

Our model is trained on one or two tasks and we report its ability to plan for every single task (Table \ref{tab_transfer_shapes}). For tasks with a moderate number of abstract states (e.g. 2 objects stacking in a $4{\times}4$ grid world has 136 abstract states in the coarsest bisimulation), our method can successfully transfer to new tasks of similar complexity without additional training. For instance, the abstract model learned from two pucks stacking can plan for placing two and three pucks in a row with a $90\%+$ success rate. The middle section of Table \ref{tab_transfer_shapes} shows tasks with coarsest abstractions at the limit of what our abstract model can represent. We can still transfer to similar tasks with a success rate higher than $75\%$. 

The bottom section of Table \ref{tab_transfer_shapes} demonstrates that our algorithm can find partial solutions even if the number of abstract states in the coarsest bisimulation exceeds the capacity of the HMM prior. We present additional transfer experiments with a house building task in Appendix B.4.


\begin{table*}[!t]
    \centering
    \scalebox{1}{
    \begin{tabular}{ccccccccc}
        \toprule
        \textbf{Source tasks} & \textbf{2S} & \textbf{3S} & \textbf{2R} & \textbf{3R} & \textbf{2\&2S} & \textbf{3ST} & \textbf{2D} & \textbf{3D} \\
        \midrule
        2S &             $\mathbf{99.9}$ \small{$\pm 0.1$} & - & $98.6$ \small{$\pm 0.5$} & - & - & - & $\mathbf{98}$ \small{$\pm 0.7$} & - \\
        2S, 2R &         $99.2$ \small{$\pm 0.9$} & - & $\mathbf{99.9}$ \small{$\pm 0.1$} & - & - & - & $81.5$ \small{$\pm 3.1$} & - \\
        \hline
        3S &             $90$ \small{$\pm 2.6$} & $\mathbf{98.2}$ \small{$\pm 0.8$} & $75.7$ \small{$\pm 2.7$} & $40.8$ \small{$\pm 14.7$} & - & $61.9$ \small{$\pm 7.5$} & $67$ \small{$\pm 3.6$} & $24.9$ \small{$\pm 7.7$} \\
        3ST &             $74.4$ \small{$\pm 3.5$} & $21.8$ \small{$\pm 6.4$} & $\mathbf{98.8}$ \small{$\pm 0.2$} & $73.9$ \small{$\pm 4.4$} & - & $\mathbf{98.8}$ \small{$\pm 0.4$} & $83.6$ \small{$\pm 2.7$} & $39.3$ \small{$\pm 4.2$} \\
        3S, 3R &         $93.8$ \small{$\pm 2.2$} & $88.8$ \small{$\pm 3.3$} & $91.5$ \small{$\pm 1.5$} & $\mathbf{88.4}$ \small{$\pm 2.7$} & - & $74.8$ \small{$\pm 6.1$} & $\mathbf{86}$ \small{$\pm 3.4$} & $\mathbf{65.2}$ \small{$\pm 6.6$} \\
        3S, 3ST &         $\mathbf{98.1}$ \small{$\pm 1.2$} & $97.8$ \small{$\pm 1.2$} & $98.1$ \small{$\pm 1.7$} & $75.2$ \small{$\pm 4.2$} & - & $92.2$ \small{$\pm 1.8$} & $84.1$ \small{$\pm 4.3$} & $51.3$ \small{$\pm 6.9$} \\
        \hline
        2\&2S &          $76.9$ \small{$\pm 8.2$} & $16.2$ \small{$\pm 2.5$} & $65.8$ \small{$\pm 3.6$} & $24.9$ \small{$\pm 4$} & $\mathbf{46.2}$ \small{$\pm 7.1$} & $4.7$ \small{$\pm 2.4$} & $46.6$ \small{$\pm 5.5$} & $12.2$ \small{$\pm 2.8$} \\
        2\&2S, 3S &      $\mathbf{92.8}$ \small{$\pm 2.7$} & $\mathbf{33.9}$ \small{$\pm 3.7$} & $67.6$ \small{$\pm 4.4$} & $30.8$ \small{$\pm 3.2$} & $38$ \small{$\pm 1.7$} & $9.6$ \small{$\pm 2.6$} & $51.5$ \small{$\pm 5.1$} & $\mathbf{16.8}$ \small{$\pm 3.4$} \\
        2\&2S, 3R &      $61.8$ \small{$\pm 5.4$} & $18.4$ \small{$\pm 3.1$} & $71.6$ \small{$\pm 3.1$} & $\mathbf{70.7}$ \small{$\pm 4.8$} & $33.9$ \small{$\pm 7.6$} & $10.3$ \small{$\pm 4.2$} & $50.4$ \small{$\pm 3$} & $10.1$ \small{$\pm 1.1$} \\
        2\&2S, 3ST &      $71.5$ \small{$\pm 7.6$} & $24.4$ \small{$\pm 3.6$} & $\mathbf{75.6}$ \small{$\pm 4.4$} & $29.6$ \small{$\pm 2.1$} & $36.1$ \small{$\pm 4.4$} & $\mathbf{33.7}$ \small{$\pm 4.5$} & $\mathbf{53.4}$ \small{$\pm 4.9$} & $15$ \small{$\pm 2.4$} \\
        \midrule
        Random Policy & $9.9$ & $0.5$ & $19.4$ & $1.9$ & $1.2$ & $3.6$ & $15.4$ & $1.5$ \\
        \bottomrule
    \end{tabular}
    }
    \caption{Transfer experiments in the Shapes World environment. In the same order as the examples in Figure \ref{fig_goals}, the tasks are stacking two/three objects (2S/3S), two/three objects in a row (2R/3R), two/three objects diagonal (2D/3D), two and two stacks (2\&2S) and stairs from three objects (3ST). We train our model with the HMM prior on one or more source tasks and then use the abstract MDP induced by the HMM prior to plan for every task. We report the \textit{success rate of reaching each goal (\%)} with a budget of 20 time steps. We trained each model 10 times over the same dataset; we report means and standard deviations.}
    \label{tab_transfer_shapes}
\end{table*}

\begin{table}[t!]
    \centering
    \scalebox{0.85}{
    \begin{tabular}{ccccc}
        \toprule
        \textbf{Game} & \textbf{DQN} & \textbf{Mean Q} & \textbf{VI} & \textbf{Random} \\
        \midrule
        Breakout        & $14$ & $19.08$ \small{$\pm 11$} & $0.83$ \small{$\pm 0.39$} & $0.66$  \\
        Space Invaders  & $55$ & $20.52$ \small{$\pm 2.95$} & $5.81$ \small{$\pm 3.12$} & $3.06$  \\
        Freeway         & $54$ & $36.21$ \small{$\pm 8.08$} & $34.95$ \small{$\pm 8.46$} & $0.2$  \\
        Asterix         & $20$ & $0.53$ \small{$\pm 0.1$} & $0.49$ \small{$\pm 0.08$} & $0.5$  \\
        \bottomrule
    \end{tabular}
    }
    \caption{DQN and abstract policies tested on MinAtar games. We train a DQN once for each game and report the mean return over the last 100 episodes. Then, we learn an abstraction for each game $10$ times and report the mean returns and standard deviations for planning in the abstract MDP with Value Iteration (VI) or averaging predicted state-action values for each abstract state (Mean Q).} 
    \label{tab_minatar}
\end{table}

\subsection{MinAtar}
\label{exp_minatar}

The challenge of the Shapes World (Subsection \ref{exp_shapes}) is that the coarsest bisimulation can have thousands of abstract states, but each task can be solved in less than ten time steps. The simplified Atari games of MinAtar pose an interesting challenge because each episode could potentially last tens or hundreds of time steps \cite{young19}. MinAtar has five Atari games--Breakout, Space Invaders, Freeway, Asterix and Seaquest\footnote{We skip Seaquest because we had trouble running it.}. The state of the games is fully observable and the dynamics are simplified. We use the same process of training a deep Q-network to create a dataset of transitions and then training our model on it, see Appendix B.5.

We test the quality of the learned abstraction in two ways. First, we employ Value Iteration in the learned abstract model to plan for the optimal policy. Planning in this domain might be challenging, as we do not use any temporal abstractions. We also test our abstraction from the perspective of compression: we average over the values of each state-action pair (predicted by the deep Q-network) belonging to each abstract state. This gives us a single value for each abstract-state action pair--we call this approach Mean Q. Intuitively, we compress the policy represented by the deep Q-network into a discrete representation.

Mean Q outperforms DQN in Breakout--an unexpected result--and reaches around $35\%$ of the performance of DQN in Space Invaders and around $60\%$ in Freeway (Table \ref{tab_minatar}). Both of the Breakout policies suffer from a high variance of returns in-between episodes; we hypothesize that the compression makes the policy more robust. Figure 2 in Appendix C.2 further analyses Mean Q on Breakout. Value Iteration can only find a useful policy for Freeway.

\section{Related Work}
\label{related_work}


Theoretical characterization of the topology of the space of abstraction, including bisimulation, was provided by \citet{li06}. Further analysis by \citet{abel2016} proved bounds on the regret of planning an optimal policy in an abstracted MDP compared to the ground MDP. A work more closely related to ours uses a deep neural net to learn the bisimulation metrics between states represented by images \cite{castro19}. The main difference between this line of work and ours is that we aim to learn an abstract MDP with discrete states, in which we can plan efficiently, whereas bisimulation metrics are more commonly used to find similar states for the purpose of transfer of policies \cite{castro10,castro11}.

The information bottleneck method defines an objective that maximizes the predictive power of a model while minimizing the number of bits needed to encode an input \cite{tishby01}. \citet{abel19} drew a connection between information bottleneck and Rate-distortion theory for the purpose of learning state abstractions. Their Expectation-Maximization-like algorithm can compress the state space given an expert policy. The information bottleneck method has also been used to regularize policies represented by deep neural networks. \citet{goyal18} improved the generalization of goal-conditioned policies by training their agent to detect decision states--states in which the agent requires information about the current goal to act optimally. \citet{teh17} used a similar objective to distill a task-agnostic policy in the multitask Reinforcement Learning setting. \citet{strouse18} used information bottleneck to control the amount of information communicated between two agents. \citet{tishby11,rubin12} study the connections between Information Theory and Reinforcement Learning.

Several recent neural-net-based methods use discrete representation for planning. \citet{serban18} can learn a factored transition given a predefined discrete state abstraction. They focus their empirical evaluation on Natural language processing tasks. \citet{kurutach18} proposed a Generative adversarial network for learning a forward model with either continuous or discrete latent states. While they show superior performance on a rope manipulation task with continuous latent states, the discrete state representation learning and planning was only evaluated on a toy 2D navigation task. Finally, \citet{corneil18} and \citet{pol20} both learn an abstract MDP with discrete states based on ground states represented as images. \cite{corneil18} used variational inference \cite{kingma13} and the Concrete distribution reparameterization trick \cite{maddison17,jang17} to learn a state representation with binary latent vectors. Their method is superior to model-free and other model-based approaches on the VizDoom 3D navigation task. Unlike our work, they do not focus on the multi-goal planning setting. But, their method is able to quickly adapt to changes in the dynamics of the environment. Recent work by \citet{pol20} learns a forward model with a continuous state representation using a loss function based on the theory of MDP homomorphisms (a generalization of bisimulation). This work differs from our work in that it does not directly learn the discrete model--it is obtained using a heuristic that samples a large number of discrete states from encodings of observed abstract states and then prunes them.

\section{Conclusion}

In this paper, we present a new method for finding discrete state abstractions from collected image states. We derive our objective function from the information bottleneck framework and learn an abstract MDP through an HMM prior conditioned on actions. Our experiments demonstrate that our model is able to learn high-quality bisimulation partitions that contain up to 1000 abstract states. We also show that our abstractions enable transfer to goals not known during training. Finally, we report experimental results on tasks with long time horizons, showing that we can use learned abstractions to compress policies learned by a deep Q-network.

In future work, we plan to address the two main weaknesses of bisimulation: it does not leverage symmetries of the state-action space to minimize the size of the found abstraction and it does not scale with the temporal horizon of the task. The former problem can be addressed with MDP homomorphisms \cite{ravindran2004}. The time horizon problems could be solved with hierarchical Reinforcement Learning.


\section*{Acknowledgments}

Special thanks to Yea-Eun Jang for graphical design in the paper and presentation materials. The authors also thank members of the GRAIL and Helping Hands groups at Northeastern, as well as anonymous reviewers, for helpful comments on the manuscript. This work was supported by the Intel Corporation, the 3M Corporation, National  Science  Foundation  (1724257, 1724191, 1763878, 1750649, 1835309), NASA (80NSSC19K1474), startup funds from Northeastern University, the Air Force Research Laboratory (AFRL), and DARPA.

\bibliography{main.bib}

\clearpage


\section*{Supplementary material}

# Supplementary Material: Learning Discrete State Abstractions With Deep Variational Inference

## A. Theoretical analysis

**Assumptions on the MDP**

Let $M = \langle S, A, T, R, \gamma \rangle$ be a ground MDP with finite $S$ and $A$, an arbitrary $R$ and a deterministic $T$. We assume there exists a bisimulation mapping $\phi_{bisim} : S \to C$ with $K$ abstract states ($|C| = K$). $\phi_{bisim}$ does not have to be the coarsest bisimulation (i.e. the one using the lowest possible number of abstract states).

**Assumptions on the model**

Let $f : S \to \mathbb{R}^D$ be a deterministic encoder. For simplicity, we use a Hidden Markov Model with a Kronecker delta observation model parameterized by $K$ component means $C = \{c_1, c_2, ..., c_K\}$ with probability density function

$$p(z|c=k) = \mathbb{I}\left[\mu_k = z\right]. \tag{1}$$

The encoder and the Hidden Markov Model induce an abstraction mapping $\phi : S \to C$

$$\phi(s) = \arg\max_{c \in C} p(c|f(s)). \tag{2}$$

We assume $\forall_{c,c' \in C} \mu_c \neq \mu_{c'}$ (i.e. no two components share their means) and $\forall_{s \in S} \exists_{c \in C} f(s) = \mu_c$ (i.e. each encoding equals to some cluster mean). The component prior $p(c)$ assign a uniform probability to each cluster to match our experimental setting.

Finally, we assume the decoder $p(y|z, a)$ is tied to the cluster assignment $\phi$. That is, each components has a parameter $q_c \in \mathbb{R}^{|A|}$, which stores the state-action value of this component for each action. The decoder probability density is a normal distribution with a mean of $q_{c,a}$ given an action $a$ and a fixed standard deviation $\sigma$

$$p(y|z, a) = \mathcal{N}(y|q_{\phi(z),a}, \sigma^2). \tag{3}$$

We can decompose the loss function into a Q-value loss $L_Q$ and a prior loss $L_P$

$$L_Q(\theta) = \frac{1}{\sigma|S||A|} \sum_{s \in S, a \in A} \left(Q^*(s,a) - q_{\phi(f(s)),a}\right)^2, \tag{4}$$

$$L_P(\theta) = -\frac{1}{|S||A|} \sum_{s_t, s_{t+1} \in S, a \in A} T(s_t, a, s_{t+1}) \log \left(\sum_{c_t, c_{t+1} \in C} p(f(s_t)|c_t) p(f(s_{t+1})|c_{t+1}) p(c_t) p(c_{t+1}|c_t, a)\right), \tag{5}$$

$$L_{IB}(\theta) = L_Q(\theta) + \beta L_P(\theta). \tag{6}$$

Supplementary Material: Learning Discrete State Abstractions With Deep Variational Inference

**Definition 1.** An abstraction mapping $\phi$ is homogeneous if for each $s, s' \in S, a \in A, \hat{c} \in C$ $\phi(s) = \phi(s')$ implies $\sum_{\hat{s} \in \phi^{-1}(\hat{c})} T(s, a, \hat{s}) = \sum_{\hat{s} \in \phi^{-1}(\hat{c})} T(s', a, \hat{s})$.

**Lemma 1.** Let $\phi$ be an abstraction mapping that is a $Q^*$-irrelevance abstraction and is also homogeneous. Then $\phi$ is a bisimulation.

*Proof.* Fix an abstract state $c$ and an action $a$. For each $s, s' \in \phi^{-1}(c)$ we have that $Q^*(s, a) = Q^*(s', a)$ (by $Q^*$-irrelevance) and $\forall \hat{c} \in C$ $\sum_{\hat{s} \in \phi^{-1}(\hat{c})} T(s, a, \hat{s}) = \sum_{\hat{s} \in \phi^{-1}(\hat{c})} T(s', a, \hat{s})$ (by $\phi$ being homogeneous). Let us expand the state-action values of $(s, a)$ using the Bellman equation

$$Q^*(s, a) = R(s, a) + \gamma \sum_{\hat{s} \in S} T(s, a, \hat{s}) V^*(s) \tag{7}$$

$$= R(s, a) + \gamma \sum_{\hat{c} \in C} T(s, a, \hat{c}) V^*(\hat{c}) \tag{8}$$

$$= R(s, a) + \gamma x. \tag{9}$$

We can perform a similar expansion for $(s', a)$

$$Q^*(s', a) = R(s', a) + \gamma \sum_{\hat{c} \in C} T(s', a, \hat{c}) V^*(\hat{c}) \tag{10}$$

$$= R(s', a) + \gamma \sum_{\hat{c} \in C} T(s, a, \hat{c}) V^*(\hat{c}) \tag{11}$$

$$= R(s', a) + \gamma x \tag{12}$$

We write $T(s, a, \hat{c})$ as a shorthand for $\sum_{\hat{s} \in \phi^{-1}(\hat{c})} T(s, a, \hat{s})$. $T(s', a, \hat{c}) = T(s, a, \hat{c})$ holds for all abstract states if $\phi(s) = \phi(s')$ because $\phi$ is homogeneous. All states mapped to a particular abstract state have the same value (for an optimal policy) due to $Q^*$-irrelevance; hence, we can write $V^*(\hat{c})$.

Since $Q^*(s, a) = Q^*(s', a)$ and $x = x$ it follows that $R(s, a) = R(s', a)$ (i.e. $\phi$ is reward-respecting). A reward-respecting homogeneous state abstraction is a bisimulation by definition.

□

**Lemma 2.** There exist model parameters $\theta$ such that $L_Q(\theta) = 0$.

*Proof. Strategy: we can achieve zero $L_Q$ by assigning states into components such that states in each component have equal state-action values for all actions. We need to show that there are enough components to perform this assignment.*

Parameters $\theta$ induce an abstraction mapping $\phi$. Assume that $L_Q(\theta) = 0$. Then for each $c \in C, s, s' \in \phi^{-1}(c), a \in A$ we have $(Q^*(s, a) - Q^*(s', a))^2 = 0$, which implies $|Q^*(s, a) - Q^*(s', a)| = 0$. Hence, $\phi$ is a $Q^*$-irrelevance abstraction (Li et al., 2006). By Li et al. (2006), for each bisimulation abstraction there exists a $Q^*$-irrelevance abstraction that is equal in size or coarser (i.e. it uses fewer abstract states). By our assumption that we have enough components to represent a bisimulation abstraction, there are also enough components to represent a $Q^*$-irrelevance abstraction. □

**Lemma 3.** There exist model parameters $\theta$ such that $L_P(\theta) = \log K$. $L_P(\theta) = \log K$ is the global minimum.

*Proof.* Under our assumptions, we can reduce the prior loss function to

$$L_P(\theta) = -\frac{1}{|S||A|} \sum_{s_t \in S, a \in A, s_{t+1} \in S} T(s_t, a, s_{t+1}) \log \frac{p(\phi(f(s_{t+1}))|\phi(f(s_t)), a)}{K}. \tag{13}$$

The minimum of this loss function is achieved when the term $p(\phi(f(s_{t+1}))|\phi(f(s_t)), a) = 1$ for all states and actions. Since we assume the transition dynamics of the ground MDP are deterministic, this transition model is only possible if the abstraction mapping $\phi$ is homogeneous. Our model can represent such abstraction mapping because, by our assumption, it can represent a bisimulation, which is homogeneous.

□

**Theorem 1.** *There exist model parameters $\theta$ that reach the global minimum of $L_{IB}(\theta) = \beta \log K$ with $\beta > 0$. The abstraction mapping $\phi$ induced by any such model parameters is a bisimulation.*

*Proof.* Since we assume we have enough components to represent a bisimulation, we can represent a $Q^*$-irrelevance abstraction (by Lemma 2) that is also homogeneous (by Lemma 3). Any such abstraction is a bisimulation by Lemma 1. □

## B. Experimental Details

We ran all of our experiments on a machine with Intel Core i7-9700K CPU @ 3.60GHz, 64GB of RAM and two Nvidia GeForce RTX 2080 Ti graphics cards. We uploaded the source code together with this supplementary material to CMT (see *codesource.zip*).

### B.1. Abstraction purity

We include a snippet of the Python code that computes abstraction purity. We use the package numpy version 1.16.1. The inputs to the function below are probability distribution over hidden states for each sample in the validation dataset and an array of labels, one for each sample.

```python
import numpy as np

def evaluate_purity(self, cluster_probs, labels):
    """
    :param cluster_probs:     NxK matrix where N is the number of samples and K the number
                                        of components.
    :param labels:            A label for each sample.
    """

    # compute the probability of each component-label pair
    label_masses = []

    for label in np.unique(labels):
        # find all samples with a particular label and sum over them
        label_mass = np.sum(cluster_probs[labels == label], axis=0)
        label_masses.append(label_mass)

    label_masses = np.stack(label_masses)

    sizes = np.sum(cluster_probs, axis=0)
    sizes[sizes == 0.0] = 1.0

    # assign a label with the highest probability mass to each cluster
    # calculate the fraction of that mass to the mass of all other labels
    purities = np.max(label_masses, axis=0) / sizes

    # average of cluster purities weighted by cluster sizes
    mean_purity = np.sum(purities * sizes) / np.sum(sizes)

    return purities, sizes, mean_purity
```

### B.2. Columns World

The deep Q-network that is used to collect the dataset has two hidden layers of 256 neurons followed by ReLU activation functions. We train it for 40000 time steps with an $\epsilon$-greedy policy; $\epsilon$ linearly decays from 1 to 0.1 over 20000 time steps. We

Supplementary Material: Learning Discrete State Abstractions With Deep Variational Inference

| Source tasks | 2SB | 3SB | 3T | 3B | 3L | 3R |
| --- | --- | --- | --- | --- | --- | --- |
| 2SB | **99.3** ±0.6 | 0.5 ±0.2 | 38 ±2.5 | 38.5 ±2.8 | 41.5 ±6 | 40.7 ±2.6 |
| 3SB | 10 ±2.5 | 61.2 ±26.4 | 3.7 ±0.3 | 3.9 ±0.9 | 3.6 ±0.7 | 2.7 ±2.6 |
| 3T | 55.2 ±3.2 | 0.6 ±0.1 | **82.4** ±5.1 | 70.2 ±5.3 | 32 ±2.9 | 26 ±4.2 |
| 3L | 57.2 ±6.5 | 2.6 ±3.6 | 23.1 ±5.2 | 28.5 ±6.4 | **84.6** ±7.1 | 71.7 ±8.4 |
| 3SB, 3T | 36.5 ±2.9 | **74.6** ±6.6 | 68.8 ±1 | 39.4 ±5.4 | 29.7 ±4.3 | 31.3 ±3.1 |
| 3T, 3L | 64.1 ±1.9 | 6.4 ±7.6 | 78.8 ±4.5 | **75** ±5.2 | 79.7 ±3.6 | **74.3** ±4.3 |
| 3SB, 3T, 3L | 58.1 ±0.7 | 60.8 ±5.2 | 62.7 ±2.5 | 56.5 ±3.1 | 68.2 ±2.2 | 49.3 ±6.2 |

*Table 1.* Transfer experiments in the Building World domain, we report the mean *success rate (%)* over 10 runs. The setup is the same as in Table 2 in the main text. The tasks are to build a building from a block and a roof (2SB), building from two stacked blocks and a roof (3SB), 2SB with an block added from top (3T), bottom (3B), left (3L) and right (3R). See Figure 1 for examples of goal states.

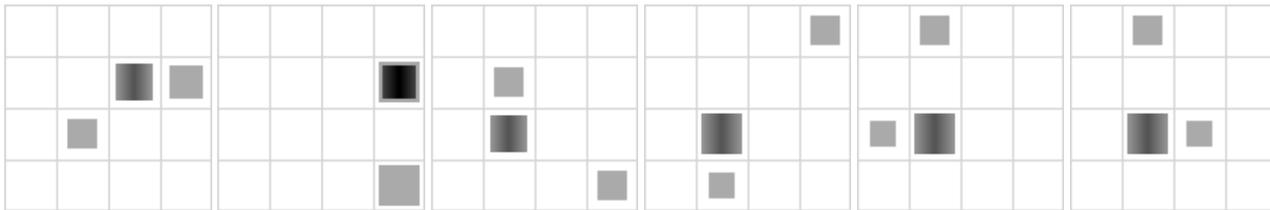

*Figure 1.* Goal states in the Building World environment. From left to right: stacking a roof on one block, stacking two blocks on top of each other and a roof on top, a roof stacked on top of a block and another block added either from the top, bottom, left or right. We show simulated depth images with inverted luminosity (the darker a pixel the higher it is); the agent does not see the grid.

use a learning rate of 0.0001, 32 mini-batch size, the target network is updated every 100 time steps and we use prioritized replay with the default settings (Schaul et al., 2016). The optimizer used for training is mini-batch gradient descent with momentum set to 0.9. The dataset for training the abstract and direct models is collected after training with $\epsilon$ set to 0.5. We compute the abstraction purity over every possible ground state.

Each state is represented as a 90-dimensional one-hot encoded vector. As a baseline, we train a model with two fully-connected layers of 128 neurons followed by ReLU activations and two heads, one for predicting the reward and the other for predicting the next state. We use a mean squared error loss for the reward prediction and cross-entropy loss for the next state (we treat each dimension of the predicted 90-dimensional vector as a probability of being in that particular state). Finally, we run an approximate partition iteration algorithm following Dean et al. (1997).

Our model with an HMM prior uses the same architecture as the above model, except it makes only one prediction: the state-action value associated with a given state-action pair. We set the number of hidden states to 6, the observation model of the HMM is 32-dimensional, encoder and model learning rates are 0.01, $\beta$ is 0.0001, the means of the HMM observations are initialized with 0 mean and 0.01 standard deviation and the diagonal covariances are initialized with -1 mean and 0.1 standard deviation before being exponentiated. We train the models using the Adam optimizer (Kingma & Ba, 2015).

### B.3. Shapes World

For dataset collection, the input image is resized to $64 \times 64$ before being fed into a deep Q-network. We use four convolutional layers with 32, 64, 128, and 256 filters; the filter size is four and the stride is set to two (each convolutional downsamples the input by a factor of two); we use "same" padding. The convolutions are followed by a single fully-connected layer with 512 neurons and a head for predicting the state-action values. The learning rate is set to 0.00005, the batch size is 32, the buffer size is 100000 and we train for 100000 steps. Actions are selected with an $\epsilon$-greedy policy–$\epsilon$ is linearly decayed from 1.0 to 0.1 over 50000 time steps. We collect a dataset of 100000 transitions after training the model with $\epsilon$ set to 0.1. 80% of the dataset is used for training and 20% for computing the abstraction purity.

Our model uses the same neural network, except we insert batch normalization between each layer and its activation function (we use ReLU) (Ioffe & Szegedy, 2015). The model predicts a 32-dimensional vector of means and a diagonal covariance, from which we sample the continuous encoding $z$. The GMM or HMM uses 1000 components (hidden states), the initial

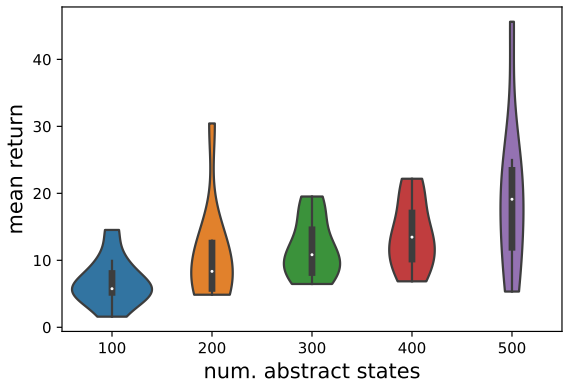

*Figure 2.* Mean return of the Mean Q agent as a function of the number of abstract states in the HMM prior in Breakout. Each setting was run 10 times and we report the results as a violin plot. We cut the plots at the minimum and maximum values of data points.

means of the components are drawn from a Gaussian distribution with 0 mean and 0.1 variance. The variances are drawn from a Gaussian with -1.0 mean and 0.1 before being exponentiated. We train the model for 50000 steps, then we collect batch normalization statistics over the whole dataset, and we resume training only the prior with a fixed encoder and unfrozen component weights $p(c_t)$ (previously held uniform fixed) for another 50000 steps. $\beta$ is set to $10^{-6}$, encoder learning rate to $10^{-3}$ and prior learning rate to $10^{-2}$. We train the model with Adam optimizer (Kingma & Ba, 2015).

To get a reward function over the abstract MDP induced by the HMM, we find abstract states with $99\%$ of ground states that are mapped to them being goal states for a given goal. We plan state-action values for each abstract-state action pair using Value Iteration and run an agent with a softmax policy with $\tau$ set to $10^{-2}$ for 100 episodes.

We ran a hyper-parameter search for the learning rates and $\beta$ on the task of stacking three pucks and then used the same parameters for all other experiments. We tried the following learning rates: $\{0.005, 0.001, 0.0005, 0.0001\}$ and $\beta$: $\{0.005, 0.001, 0.0005, 0.0001, 0.00005, 0.00001\}$.

### B.4. Stacking Buildings

The hyper-parameters are the same as for Shapes World (Section B.3).

### B.5. MinAtar

We use a deep Q-network architecture and a training script provided by Young & Tian (2019). We collect 100000 transitions with $\epsilon$ set to 0.1 after training it for 3M time steps. The authors train up to 5M time steps, but we disable sticky actions and difficulty ramping, making the games easier.

The details of our model are similar to Section B.3, except we use a smaller convolutional network with 32 and 64 filters in two layers, filter size set to three and stride set to one. We do not use batch normalization and the hidden layer after convolutions has only 128 neurons. $\beta$ is set to $5 * 10^{-5}$ and the rest of the parameters stay the same. For our abstract agent, we do not threshold goal states and set $\tau$ to $5 * 10^{-4}$.

We ran a hyper-parameter search on Breakout and then used the best parameters for all other games. The search space was the same as in Section B.2.

## C. Additional Experiments

### C.1. Stacking buildings

The set up for this experiment is the same as Shapes World (Section 6.2). We instantiate five different tasks (Figure 1) and report our transfer results in Table 1. The tasks are more difficult than the ones in Shapes World. All tasks except for the first have too many abstract states in the coarsest bisimulation to be represented by the 1000 hidden states available in the HMM

prior.

One data point of interest is our models ability to generalize between different orientations of buildings (Figure 1, images 3 (3T–top), 4 (3B–bottom), 5 (3L–left), and 6 (3R–right)). The abstraction trained on the 3T task can generalize to 3B, but not to 3L and 3R. Conversely, 3L can generalize to 3R, but not 3R and 3L. Training on 3T and 3L leads to an abstraction that can solve both 3B and 3R (Table 1 line 6), albeit not as well as the abstractions from 3T (line 3) and 3L (line 4) separately.

### C.2. MinAtar

In Figure 2, we investigate the impact of the number of abstract states on the performance of the Mean Q abstract agent (Section 6.3) in Breakout. Even though there is a high variance between the qualities of abstraction learned in different runs, the violin plot shows an approximately linear dependence between the number of abstract states and the mean return.



\end{document}